\documentclass{article}
\usepackage[T1]{fontenc}
\usepackage{geometry}
\usepackage{amsmath}
\usepackage{amssymb}
\usepackage{algorithm}
\usepackage[noend]{algpseudocode}
\usepackage{graphicx}

\DeclareMathOperator*{\argmax}{argmax}
\DeclareMathOperator*{\argmin}{argmin}
\DeclareMathOperator*{\expect}{E}
\DeclareMathOperator*{\UCB}{UCB}
\DeclareMathOperator*{\LCB}{LCB}

\title{Efficient exploration of zero-sum stochastic games}

\author{
    Carlos Martin\textsuperscript{1}\\
    \texttt{cgmartin@cs.cmu.edu}
    \and
    Tuomas Sandholm\textsuperscript{1,2,3,4}\\
    \texttt{sandholm@cs.cmu.edu}
}

\date{
    \textsuperscript{1}Computer Science Department, Carnegie Mellon University \\
    \textsuperscript{2}Optimized Markets, Inc. \\
    \textsuperscript{3}Strategic Machine, Inc. \\
    \textsuperscript{4}Strategy Robot, Inc.
}

\begin{document}

\maketitle

\begin{abstract}
We investigate the increasingly important and common game-solving setting where we do not have an explicit description of the game but only oracle access to it through gameplay, such as in financial or military simulations and computer games. During a limited-duration learning phase, the algorithm can control the actions of both players in order to try to learn the game and how to play it well. After that, the algorithm has to produce a strategy that has low exploitability. Our motivation is to quickly learn strategies that have low exploitability in situations where evaluating the payoffs of a queried strategy profile is costly. For the stochastic game setting, we propose using the distribution of state-action value functions induced by a belief distribution over possible environments. We compare the performance of various exploration strategies for this task, including generalizations of Thompson sampling and Bayes-UCB to this new setting. These two consistently outperform other strategies.
\end{abstract}

\section{Introduction}

We study the problem of how to efficiently explore zero-sum games whose payoffs and dynamics are initially unknown. The agent is given a certain number of episodes to learn as much useful information about the game as possible. During this learning, the rewards obtained in the game are fictional and thus do not count toward the evaluation of the final strategy. 
After this exploration phase, the agent must recommend a strategy that should be minimally exploitable by an adversary (who has complete knowledge of the environment and can thus play optimally against it). This setup is called {\em pure exploration} in the single-agent reinforcement learning literature.
This is an important problem for simulation-based games in which a black-box simulator is queried with strategies to obtain samples of the players' resulting utilities~\cite{Vorobeychik2009}, as opposed to the rules of the game being explicitly given. 
For example, in many military settings, war game simulators are used to generate strategies, and then the strategies need to be ready to deploy in case of actual war~\cite{Marchesi2019}. Another prevalent example is finance, where trading strategies are generated in simulation, and then they need to be ready for live trading. A third example is video games such as Dota 2~\cite{Berner2019} and Starcraft II~\cite{Vinyals2019}, where AIs can be trained largely through self-play.

This raises the challenge not only of learning approximate equilibria with noisy observations, but of learning in as few queries as possible, since running the simulator is usually an expensive operation. This is true even for benchmark computer games like Dota 2 and Starcraft II, which are costly to explore due to their length and complexity. Furthermore, unlike this paper, that prior work on Dota 2 and Starcraft II did not evaluate the exploitability of the learned strategies and did not directly target the minimization of exploitability.

In the context of single-agent reinforcement learning, Q-learning~\cite{Watkins1992} tries to learn state-action values directly in a model-free way, that is, without learning the structure of the underlying environment. Dearden et al.~\cite{Dearden1998} extended Q-learning to incorporate uncertainty by propagating probability distributions over the Q values in order to compute a myopic approximation of the value of information, which measures the expected improvement in future decision quality that can be gained from exploration. Bellemare et al.~\cite{Bellemare2017} and Donoghue et al.~\cite{Donoghue2018} argue for the importance of the Q value distribution and propose a new version of Bellman updating that incorporates uncertainty.

A similar problem has been studied in the context of single-agent model-based reinforcement learning. {\em Posterior sampling reinforcement learning}~\cite{Osband2017,Agrawal2017} samples an environment from the agent's belief distribution, follows a policy that is optimal with respect to it, and then updates the agent's beliefs about the environment with the resulting observations. 

Even in single-agent settings, sampling a new policy on every step within an episode is inefficient because it does not perform so-called {\em deep exploration}, which accounts not only for information gained by taking an action but also for how the action may position the agent to more effectively acquire information later~\cite{Russo2018}. To address this limitation, in deep exploration a single policy is chosen at the beginning of each episode and followed for its duration.
Osband et al.~\cite{Osband2016} and Osband and Van Roy~\cite{Osband2019} propose an approach to deep exploration that chooses actions that are optimal with respect to a value function that is sampled from an ensemble. Each element of the ensemble is a deep neural network trained with deep Q-learning~\cite{Mnih2015}, and the ensemble constitutes a belief distribution over possible value functions of the environment. It incentivizes experimentation with actions of uncertain value because uncertainty induces variance in the sampled value estimate.
Chen et al.~\cite{Chen2017} also use an ensemble of Q functions but, instead of sampling from them, use the resulting upper confidence bounds. Chaudhuri et al.~\cite{Mavrin2019} combine a decaying schedule with exploration bonuses computed from upper quantiles of the learned distribution.
Littman~\cite{Littman1994} describes a Q-learning-like algorithm for finding optimal policies for one player in stochastic games when playing against an opponent that the algorithm does not control. 

Sandholm and Crites~\cite{Sandholm1996} study reinforcement learning in a repeated game. They study the role of other agents making the setting stochastic for a learner, the role of exploration, and convergence to cycles of different lengths, and how recurrent neural networks can, in principle, help with those issues.  
Claus and Boutilier~\cite{Claus1998} study reinforcement learning in cooperative settings, showing that several optimistic exploration strategies increase the likelihood of reaching an optimal equilibrium.  
Wang and Sandholm~\cite{Wang2002} describe an algorithm that converges almost surely to an optimal equilibrium in any team stochastic game. 
Hu and Wellman~\cite{Hu2003} present Nash Q-learning for general-sum stochastic games, which is guaranteed to converge to an equilibrium if all agents follow the algorithm and the stage games satisfy certain highly restrictive conditions. 
Ganzfried and Sandholm~\cite{Ganzfried2009} design algorithms for computing equilibria in special classes of stochastic games of imperfect information. 
Casgrain et al.~\cite{Casgrain2019} study Nash Q-learning but they use a neural network to model the Q function, decomposing it into a sum of the state value function and a specific form of action advantage function. Sokota et al.~\cite{Sokota2019} use neural networks to learn a mapping from mixed-strategy profiles to deviation payoffs in order to approximate role-symmetric equilibria in large simulation-based games.

In this paper we study model-driven exploration for two-player zero-sum normal-form games and stochastic games (of the standard kind where the agents know the state but they make parallel moves in each state, which begets a form of imperfect information). The exploring agent controls \textit{both} players and tries to quickly learn a minimally-exploitable strategy. 

\section{Normal-form games}

A zero-sum normal-form game is a function \(u : A_1 \times A_2 \rightarrow \mathbb{R}\) where \(A_1\) is a set of actions available to Player 1 and \(A_2\) is a set of actions available to Player 2. Both players choose actions simultaneously. If they jointly play the action profile \((a_1, a_2)\), the mean payoff obtained by Player 1 is \(u(a_1, a_2)\) and the mean payoff obtained by Player 2 is \(-u(a_1, a_2)\).

Let \(\triangle S \subseteq S \rightarrow \mathbb{R}\) be the set of all probability distributions on \(S\). Then \(\triangle A_1\) and \(\triangle A_2\) are the sets of mixed strategies available to each player. The maxmin strategy of Player 1, which maximizes their minimum payoff, is
\begin{equation}
    \sigma_1^* = \argmax_{\sigma_1 \in \triangle A_1} \min_{\sigma_2 \in \triangle A_2} u(\sigma_1, \sigma_2) \ \mbox{ where }  
\end{equation}
\begin{equation}
    u(\sigma_1, \sigma_2) = \sum_{a_1 \in A_1} \sum_{a_2 \in A_2} u(a_1, a_2) \sigma_1(a_1) \sigma_2(a_2)
\end{equation}

Any fixed strategy for Player 1 has a deterministic best response from Player 2. Hence minimization over the infinite set of mixed strategies inside the argmax can be turned into a minimization over the finite set of pure strategies:
\begin{equation}
    \sigma_1^* = \argmax_{\sigma_1 \in \triangle A_1} \min_{a_2 \in A_2} u(\sigma_1, a_2)
\end{equation}
where the summation is now carried out over \(A_1\) alone. Similarly, the minmax strategy of Player 2 is
\begin{equation}
    \sigma_2^* = \argmin_{\sigma_2 \in \triangle A_2} \max_{\sigma_1 \in \triangle A_1} u(\sigma_1, \sigma_2)
\end{equation}

By the celebrated minmax theorem \cite{Neumann1928}, the maxmin and minmax payoffs are equal. That quantity is called the value of the game.

\subsection{Playing optimally under uncertainty}

Suppose Player 1 is uncertain about the payoffs \(u\). More precisely, her beliefs about \(u\) are described by some distribution \(U\). To minimize the expected exploitability of her strategy---that is, to maximize the expected minimum payoff of such a strategy---she should play
\begin{equation}
    \argmax_{\sigma_1 \in \triangle A_1} \expect_{u \sim U} \min_{\sigma_2 \in \triangle A_2} u(\sigma_1, \sigma_2)
\end{equation}
We coin this the maxmeanmin strategy of Player 1. Similarly, we will call
\begin{equation}
    \argmin_{\sigma_2 \in \triangle A_1} \expect_{u \sim U} \max_{\sigma_1 \in \triangle A_2} u(\sigma_1, \sigma_2)
\end{equation}
the minmeanmax strategy of Player 2, assuming their beliefs about \(u\) are described by the distribution \(U\). These definitions would work even if the belief distributions \(U\) of the two players were different, but in our setting they are the same because we are controlling the exploration by both players.

\subsection{Solutions via linear programming}

The maxmin strategy is the solution to the following linear program over the variables \(v \in \mathbb{R}\) and \(\sigma_1 \in \mathbb{R}^{|A_1|}\):
\begin{equation}
\begin{split}
    \text{maximize} \quad & v \\
    \text{subject to} \quad & v \leq u(\sigma_1, a_2) \quad \forall a_2 \in A_2 \\
    & \sigma_1 \geq \mathbf{0} \\
    & \mathbf{1} \cdot \sigma_1 = 1
\end{split}
\end{equation}
where \(\mathbf{0}\) is the vector of zeros and \(\mathbf{1}\) is the vector of ones. The dual variables of the solution contain the minmax strategy of Player 2. We observe that a similar approach can be used to solve for the maxmeanmin strategy. Specifically, we can use a Monte Carlo estimate of the expectation:
\begin{equation}
\begin{split}
    \argmax_{\sigma_1 \in \triangle A_1} \frac{1}{K} \sum_{k = 1}^K \min_{\sigma_2 \in \triangle A_2} u_k(\sigma_1, \sigma_2)
\end{split}
\end{equation}
where \(u_k \sim U\) are payoff functions sampled from the belief distribution. 
The maxmeanmin strategy problem can then be formulated as a linear program over the variables \(v \in \mathbb{R}^K\) and \(\sigma_1 \in \mathbb{R}^{|A_1|}\):
\begin{equation}
\begin{split}
    \text{maximize} \quad & \mathbf{1} \cdot v \\
    \text{subject to} \quad & v_k \leq u_k(\sigma_1, a_2) \quad \forall a_2 \in A_2, k \in [K] \\
    & \sigma_1 \geq \mathbf{0} \\
    & \mathbf{1} \cdot \sigma_1 = 1
\end{split}
\end{equation}
Analogous expressions hold for the minmax and minmeanmax strategies of Player 2.

\subsection{Connection to multi-armed bandits}

In a multi-armed bandit problem, an agent faces a set of actions with uncertain payoff distributions from which they can sample a limited number of times. In the standard version of that problem, the agent must---over a given period of time---acquire information about the mean payoff of each action while simultaneously trying to maximize the cumulative payoff. Because of this tradeoff, multi-armed bandit problems exemplify the kind of exploration-exploitation dilemma that is central to reinforcement learning.

In the \textit{pure exploration} version of that problem, there is a learning phase first, during which the rewards obtained are fictional and do not count toward the evaluation~\cite{Bubeck2009}. Then the agent recommends an arm (i.e., action) to play going forward. The agent's performance is measured purely by the effectiveness of her recommended action. This performance measure is called  \textit{simple regret}, in contrast to the cumulative regret of the standard problem where rewards throughout the process count. Our paper focuses on the multiagent generalization of pure exploration.

Garivier et al.~\cite{Garivier2016} study the problem of pure exploration in the context of a \textit{sequential}-move game with the aim of identifying an \(\varepsilon\)-maxmin action with probability at least \(1 - \delta\). Marchesi et al.~\cite{Marchesi2019} tackle the problem of learning equilibria in simulation-based games of infinite strategy spaces with high confidence using as few simulator queries as possible. They propose an algorithm for the fixed-confidence setting (guaranteeing the desired confidence level while minimizing the number of queries) and one for the fixed-budget setting (maximizing the confidence without exceeding the given maximum number of queries).

We study the following version of this problem: An agent faces a normal-form (that is, \textit{simultaneous}-move) game with unknown mean payoffs for each action profile. The agent has a belief distribution \(U : \triangle (A_1 \times A_2 \rightarrow \mathbb{R})\) that describes her beliefs about the mean payoff of each action profile.

In each of \(T\) episodes, the agent chooses an action profile (that is, actions for both players), observes a sample from the payoff distribution for that action profile, and updates her belief distribution accordingly. After the \(T\) episodes are over, the agent recommends a strategy \(\hat{\sigma}_1\) for Player 1. The simple regret \(R(\hat{\sigma}_1)\) of this recommended strategy is
\begin{equation}
    \max_{\sigma_1 \in \triangle A_1} \min_{\sigma_2 \in \triangle A_2} u(\sigma_1, \sigma_2) - \min_{\sigma_2 \in \triangle A_2} u(\hat{\sigma}_1, \sigma_2)
\end{equation}
It measures how exploitable the recommended strategy is in comparison to the true maxmin strategy. The original multi-armed bandit problem is a single-player version of this problem, that is, one where \(|A_2| = 1\).

Since the agent is trying to minimize the expectation of this quantity, if her beliefs about the mean payoffs are modelled by \(U\), she should recommend the strategy that is maxmeanmin under \(U\). This leaves the question of how she should explore over the \(T\) episodes, since this determines how useful the belief distribution she ends up with is.

\subsection{Exploration strategies}

The simplest exploration strategy, which we call the \textbf{random strategy}, selects an action profile \((a_1, a_2) \in A_1 \times A_2\) uniformly at random on every episode:
\begin{equation}
\begin{split}
    (a_1, a_2) &\sim \text{uniform}(A_1 \times A_2)
\end{split}
\end{equation}

Even a simple exploration strategy like this is guaranteed to converge to correct beliefs about \(u\)---and hence to recommending the optimal strategy---because every action profile is explored an infinite number of times. It may, however, converge significantly more slowly than other methods.

Another strategy---called the round robin, least-explored, or \textbf{min count} strategy---selects uniformly at random from the action profiles that have been explored the least:
\begin{equation}
\begin{split}
    (a_1, a_2) &\sim \text{uniform}\left(\argmin_{(a_1, a_2) \in A_1 \times A_2} n(a_1, a_2)\right)
\end{split}
\end{equation}
where \(n(a_1, a_2)\) denotes the number of times \((a_1, a_2)\) has been explored. It allocates an equal (or almost equal) number of rounds to every action profile. Like the random strategy, it is guaranteed to converge to correct beliefs.

Our next exploration strategy, which we call the \textbf{greedy strategy}, samples an action profile from the distribution over action profiles created by the maxmeanmin and minmeanmax strategies for players 1 and 2, respectively:
\begin{equation}
\begin{split}
    \sigma_1 &= \argmax_{\sigma_1 \in \triangle A_1} \expect_{u \sim U} \min_{\sigma_2 \in \triangle A_2} u(\sigma_1, \sigma_2) \\
    \sigma_2 &= \argmin_{\sigma_2 \in \triangle A_2} \expect_{u \sim U} \max_{\sigma_1 \in \triangle A_2} u(\sigma_1, \sigma_2)
\end{split}
\end{equation}

The intuition behind adopting such an exploration strategy (possibly in combination with other strategies, as we will see below) is that, if the explorer is reasonably certain about where the Nash equilibrium of the game is, it makes sense to try to refine her knowledge about the payoffs near this equilibrium rather than waste time on action profiles and strategies that are thought to be far from equilibrium.

The greedy strategy alone is not guaranteed to converge to correct beliefs because it may get stuck with false maxmeanmin and minmeanmax strategies from which it samples forever, ignoring the other action profiles. To resolve this problem, one can combine it with the random strategy to produce the so-called \textbf{\(\varepsilon\)-greedy strategy}. In every episode, it follows the random strategy with probability \(\varepsilon\) and the greedy strategy with probability \(1-\varepsilon\).

Our next exploration strategy is based on Thompson sampling, a remarkably simple but successful heuristic for solving exploration-exploitation dilemmas in multi-armed bandit problems that consists of playing an action according to the probability it is the optimal action. This is done by sampling a belief from the belief distribution and then acting optimally with respect to that belief.
Originally introduced by Thompson~\cite{Thompson1933}, it was rediscovered and analyzed in the context of the multi-armed bandit problem~\cite{Ortega2010}. In recent years, it has become a popular approach for reinforcement learning~\cite{Russo2018}, and convergence results have been obtained that show it is asymptotically optimal and well-behaved~\cite{Osband2017,Agrawal2017}.
Our adaptation of this approach to games, which we call the \textbf{Thompson strategy}, samples from the Nash equilibrium of a payoff matrix \(u\), which is in turn sampled from the belief distribution \(U\). Seen another way, it selects an action profile according to the probability it would be played under the true (unknown) game by optimal players:
\begin{equation}
\begin{split}
    \sigma_1 &= \argmax_{\sigma_1 \in \triangle A_1} \min_{\sigma_2 \in \triangle A_2} u(\sigma_1, \sigma_2) \\
    \sigma_2 &= \argmin_{\sigma_2 \in \triangle A_2} \max_{\sigma_1 \in \triangle A_1} u(\sigma_1, \sigma_2)
\end{split}
\end{equation}
where \(u \sim U\). Like the greedy exploration strategy, it biases sampling toward what is believed to be the Nash equilibrium, but unlike greedy exploration, it also leaves room for exploration according to how uncertain the agent is. The more certain the agent is of the true payoffs \(u\), and thus of where the Nash equilibrium lies, the more often it will sample action profiles that are played by that equilibrium, refining its knowledge of the equilibrium.

Another exploration strategy in single-agent settings is called the \textbf{UCB strategy}. It is based on an approach to the multi-armed bandit problem with the same name. This strategy, also known as ``optimism in the face of uncertainty'', selects the action with the highest \textit{upper confidence bound} on its mean payoff.
Our adaptation of this exploration strategy to the setting of a normal-form game samples an action profile from the action profile distribution induced by the following pair of strategies:
\begin{equation} \label{eq:ucb}
\begin{split}
    \sigma_1 &= \argmax_{\sigma_1 \in \triangle A_1} \UCB_{u \sim U} \min_{\sigma_2 \in \triangle A_2} u(\sigma_1, \sigma_2) \\
    \sigma_2 &= \argmin_{\sigma_2 \in \triangle A_2} \LCB_{u \sim U} \max_{\sigma_1 \in \triangle A_1} u(\sigma_1, \sigma_2)
\end{split}
\end{equation}
where UCB and LCB are episode-dependent upper and lower confidence bounds, respectively. Note that each strategy is optimistic from the perspective of its corresponding player. We study two instantiations of UCB, which we will present in the next two paragraphs, respectively.

Auer et al.~\cite{Auer2002} introduced the \textbf{UCB1} algorithm for single-agent settings and proved it achieves optimal regret up to a multiplicative constant. It uses the following upper-confidence bound for utilities bounded by the interval \([0,1]\):
\begin{equation}
    \UCB_{u \sim U} u(a) = \expect_{u \sim U} u(a) + \sqrt{\frac{2 \ln n}{n(a)}}
\end{equation}
where \(n(a)\) is the number of times \(a\) has been explored and \(n = \sum_a n(a)\) is the total number of explorations. 
To adapt this to games, we have to define some analog of action counts for mixed strategies. We do this as follows:
\begin{equation}
\begin{split}
    n(\sigma_1) &= \sum_{a \in A_1} \sigma_1(a_1) n(a_1) \\
    u(\sigma_1) &= \min_{\sigma_2 \in \triangle A_2} u(\sigma_1, \sigma_2) 
\end{split}
\end{equation}

Kaufmann et al.~\cite{Kaufmann2012} introduced a Bayesian version of the UCB strategy called \textbf{Bayes-UCB} and proved it satisfies finite-time regret bounds that imply asymptotic optimality. It selects the action whose mean payoff \(1 - \frac{1}{n}\) quantile is highest. Our adaption of the Bayes-UCB strategy to this normal-form game setting is as follows. Sample \(K\) payoff matrices \(u_k \sim U\) from the belief distribution. Then compute
\begin{equation}
\begin{split}
    \sigma_1 &= \argmax_{\sigma_1 \in \triangle A_1} \operatorname*{quantile}_{k \in [K]}\left(\min_{a_2 \in A_2} u_k(\sigma_1, a_2), 1-\frac{1}{n}\right)
\end{split}
\end{equation}
\begin{equation}
\begin{split}
    \sigma_2 &= \argmin_{\sigma_2 \in \triangle A_2} \operatorname*{quantile}_{k \in [K]}\left(\max_{a_1 \in A_1} u_k(a_1, \sigma_2), 0+\frac{1}{n}\right) \\
\end{split}
\end{equation}
where \(\operatorname*{quantile}_{k \in [K]}(a_k, q)\) is the \(q\)th (linearly-interpolated) empirical quantile of the data \(a_k\). 
The optimization technique used to find \(\sigma_1\) and \(\sigma_2\) may vary. 
In our experiments, we simply optimize over finite sets of strategies that are sampled from \(\triangle A_1\) and \(\triangle A_2\), respectively. This finite-sample empirical Bayes-UCB can be seen as an interpolation of the true Bayes-UCB (with infinite samples) and Thompson sampling (with a single sample).

\section{Stochastic games}

We now turn to stochastic games, which generalize repeated normal-form games to multiple states and Markov decision processes (MDPs) to multiple agents.
Stochastic games were introduced by Shapley~\cite{Shapley1953} to model a game played in a sequence of steps. At the beginning of each step the game is in some state. Each player simultaneously selects an action and receives a payoff that depends on the current state and the joint action profile. The game then transitions to a new state---possibly stochastically---that depends on the current state and the joint action profile. This process continues either forever or until a fixed number of steps.

Here we focus on the case of a two-player zero-sum discounted infinite-horizon stochastic game. It is defined as a tuple \((S, A_1, A_2, \gamma, R, T)\) where \(S\) is a set of states, \(A_1\) is a set of actions for Player 1, \(A_2\) is a set of actions for Player 2, \(\gamma \in [0, 1]\) is a discount factor, \(R : S \rightarrow (A_1 \times A_2 \rightarrow \mathbb{R})\) gives the mean reward for a state and action profile (when actually playing we only get a sample), and \(T : S \rightarrow (A_1 \times A_2 \rightarrow \triangle S)\) is a state transition function that yields a distribution of next states for a state and action profile. Stochastic games subsume simultaneous-move extensive-form perfect-information games by letting \(\gamma = 1\) and adding transitions from all terminal states to a zero-reward absorbing state.

A policy for Player 1 is an assignment of a strategy to every state ($\pi_1 : S \rightarrow \triangle A_1$) and likewise for Player 2 ($\pi_2 : S \rightarrow \triangle A_2$).
The (expected) cumulative reward obtained by such a pair of policies when starting from state \(s_0 \in S\) is
\begin{equation}
\begin{split}
    V_{\pi_1,\pi_2}(s_0)
    &= \expect \sum_{t=0}^\infty \gamma^t R(s_t)(x_t, y_t)
\end{split}
\end{equation}
where \(x_t \sim \pi_1(s_t), y_t \sim \pi_2(s_t), s_{t+1} \sim T(s_t)(x_t, y_t)\). The goal of Player 1 is to maximize the cumulative reward, while the goal of Player 2 is to minimize it. Assuming both players play optimally, from the perspective of Player 1, the value of a state and action profile is \(Q : S \rightarrow (A_1 \times A_2 \rightarrow \mathbb{R})\),
\begin{equation} 
    Q(s)(a_1,a_2) = R(s)(a_1,a_2) + \gamma \langle T(s)(a_1,a_2), V \rangle
\label{eq:valiter}
\end{equation}
where \(\langle f, g \rangle = \int_{s \in S} f(s) g(s) \,\mathrm{d}s\) is the inner product of functions over \(S\) (with the integral replaced by a sum in a discrete state space) and \(V : S \rightarrow \mathbb{R}\),
\begin{equation}
    V(s) = \max_{\sigma_1 \in \triangle A_1} \min_{\sigma_2 \in \triangle A_2} Q(s)(\sigma_1, \sigma_2)
\end{equation}
is the value of a state. For any state \(s\), \(Q(s)\) is a normal-form game and \(V(s)\) is its value.

\subsection{Solving stochastic games}

Several algorithms for solving stochastic games have been developed. The first is the \textbf{Shapley algorithm}~\cite{Shapley1953}. It consists of repeatedly applying the Bellman operator to a value function \(V : S \rightarrow \mathbb{R}\) until the value function converges to a fixed point, much like value iteration for single-agent MDPs. 
See Algorithm \ref{alg:shapley}.
\begin{algorithm}
    \caption{Shapley algorithm}
    \label{alg:shapley}
    \begin{algorithmic}
        \Function{Shapley}{$R$, $T$, $\gamma$}
            \State \(V \gets \mathbf{0}\)
            \Loop
                \State \(Q \gets R + \gamma \langle T, V \rangle\)
                \State \(V_\text{new} \gets \Call{solveNFGs}{Q}.\text{values}\)
                \If{$\|V - V_\text{new}\|_\infty < \varepsilon$}
                    \State \Return \(Q\)
                \EndIf
                \State \(V \gets V_\text{new}\)
            \EndLoop
        \EndFunction
    \end{algorithmic}
\end{algorithm}

The auxiliary function \textsc{solveNFGs} uses linear programming to solve a batch of normal-form games. It returns their values, maxmin strategies, and minmax strategies. The Shapley algorithm converges slowly if \(\gamma \approx 1\). 

The \textbf{Hoffman-Karp algorithm}~\cite{Hoffman1966}---which we present as Algorithm~\ref{alg:hoffman}---converges in fewer iterations.
\begin{algorithm}
    \caption{Hoffman-Karp algorithm}
    \begin{algorithmic}
        \Function{HoffmanKarp}{$R$, $T$, $\gamma$}
            \State \(V \gets \mathbf{0}\)
            \Loop
                \State \(Q \gets R + \gamma \langle T, V \rangle\)
                \State \(\pi_2 \gets \Call{solveNFGs}{Q}.\text{minmax-strategies}\)
                \State \(R_{\pi_2} \gets \) reward function induced by \(R\) under \(\pi_2\)
                \State \(T_{\pi_2} \gets \) transition function induced by \(T\) under \(\pi_2\)
                \State \(V_\text{new} \gets \Call{solveMDP}{R_{\pi_2}, T_{\pi_2}, \gamma}.\text{max}\)
                \If{$\|V - V_\text{new}\|_\infty < \varepsilon$}
                    \State \Return \(Q\)
                \EndIf
                \State \(V \gets V_\text{new}\)
            \EndLoop
        \EndFunction
    \end{algorithmic}
\label{alg:hoffman}
\end{algorithm}
In the inner loop, it solves the MDP induced by the policy \(\pi_2\) for Player 2. Specifically, in the algorithm, 
\begin{equation}
\begin{split}
    R_{\pi_2}(s)(a) &= \sum_{a_2 \in A_2} \pi_2(a_2) R(s)(a, a_2) \\
    T_{\pi_2}(s)(a) &= \sum_{a_2 \in A_2} \pi_2(a_2) T(s)(a, a_2)
\end{split}
\end{equation}
To solve the corresponding MDP, we can use any of the standard algorithms for solving MDPs, such as policy iteration (Algorithm \ref{alg:politer}), which improves a policy with respect to its own value function until convergence.
\begin{algorithm}
    \caption{Policy iteration algorithm}
    \label{alg:politer}
    \begin{algorithmic}
        \Function{PolicyIteration}{$R$, $T$, $\gamma$}
            \State \(\pi \gets \mathbf{0}\)
            \Loop
                \State \(R_\pi \gets \) reward function induced by \(R\) under \(\pi\)
                \State \(T_\pi \gets \) transition function induced by \(T\) under \(\pi\)
                \State \(V \gets (I - \gamma T_\pi)^{-1} R_\pi\)
                \State \(Q \gets R + \gamma \langle T, V \rangle\)
                \State \(\pi_\text{new} \gets Q.\text{argmax}\)
                \If{$\pi = \pi_\text{new}$}
                    \State \Return Q
                \EndIf
                \State \(\pi \gets \pi_\text{new}\)
            \EndLoop
        \EndFunction
    \end{algorithmic}
\end{algorithm}

\subsection{Distributional perspective}

As in the normal-form game setting discussed earlier, suppose an agent is trying to learn minimally-exploitable strategies for a two-player zero-sum stochastic game with unknown rewards and transitions. Her beliefs about the game structure are described by a joint distribution over reward and transition functions.
Her goal is to find a least-exploitable policy for Player 1. More precisely, let
\begin{equation}
    \mathrm{minpay}(\pi_1) = \min_{\pi_2 : S \rightarrow \triangle A_2} \frac{1}{|S|} \sum_{s \in S} V_{\pi_1,\pi_2}(s)
\end{equation}

The goal of the agent is to recommend, at the end of the experiment, a Player 1 policy \(\hat{\pi}_1\) that minimizes the regret
\begin{equation}
    R(\hat{\pi}_1) = \max_{\pi_1 : S \rightarrow \triangle A_1} \mathrm{minpay}(\pi_1) - \mathrm{minpay}(\hat{\pi}_1)
\end{equation}

In the first \(T\) episodes, the agent selects actions for both players to explore the game. Recall that, for any given state \(s\), the matrix of Q values \(Q(s)\) constitutes a normal-form game. This means that, once we have a distribution over Q matrices for a particular state, we can treat it as a normal-form game with uncertainty. Hence if the agent's beliefs about the environment's Q values are captured by a belief distribution \(U\), the optimal policy to recommend is
\begin{equation}
    \hat{\pi}_1(s) = \argmax_{\sigma_1 \in \triangle A_1} \expect_{Q \sim U} \min_{\sigma_2 \in \triangle A_2} Q(s)(\sigma_1, \sigma_2)
\end{equation}

\subsection{Exploration strategies}

We use the same exploration strategies that we introduced for normal-form games, except that the mean payoff matrices are now given by a Q function evaluated at the desired state. The belief distribution over Q functions is approximated by sampling \(K\) stochastic games (that is, reward and transition function pairs) from the belief distribution over stochastic games and then solving them with any of the algorithms used to solve stochastic games. At the end of each episode, we update our beliefs about the stochastic game in accordance with the observed rewards and transitions. (As discussed in the introduction, we do not update beliefs after every time step because that would not achieve deep exploration.)

\section{Experiments}

We empirically compare the performance of the exploration strategies we proposed. We measure how quickly the exploitability of the recommended final strategy decreases over the course of episodes in the learning phase.
We benchmark on discrete finite-state environments. The next-state distributions of each state-action profile are categorical, while the reward distributions of each state-action profile are Bernoulli (binary).
In the experiments we assume that our initial beliefs about the parameters of these distributions are uninformatve. Specifically, they are Jeffreys priors~\cite{Jeffreys1946}, which are parameterization-independent. Thus, for any given state and action profile, our beliefs about its reward distribution are modeled by a beta distribution, while our initial beliefs about its next-state distribution are modelled by a Dirichlet distribution. These distributions have the advantage that they are conjugate priors and are very easy to update under new observations (by simply incrementing the concentration parameter corresponding to the observed outcome).

It is not necessary to start with such beliefs. For example, we may be certain there are only two possible environments, in which case the agent's beliefs are modelled by a mixture of the two. Unlike the Jeffreys prior case, the reward and transition distributions become highly correlated. Nonetheless, our exploration strategies are independent of the details about the environment belief distribution, since they only require the ability to sample from it.

The experimental parameters to be considered in the case of normal-form games are the number of actions \(|A_1|\) and \(|A_2|\) for each player, the number of samples used to approximate the distribution of payoffs, the number of trials for each method, the total number of episodes \(T\), and the number of strategies sampled from the strategy spaces of each player that are used to estimate the UCBs and LCBs of the UCB exploration strategy (we use 100 for each player). In the case of a stochastic game we also have the discount factor \(\gamma\), the number of states \(|S|\), and the number of steps per episode.

\begin{figure}[!ht]
\centering
\noindent \includegraphics[width=0.8\columnwidth]{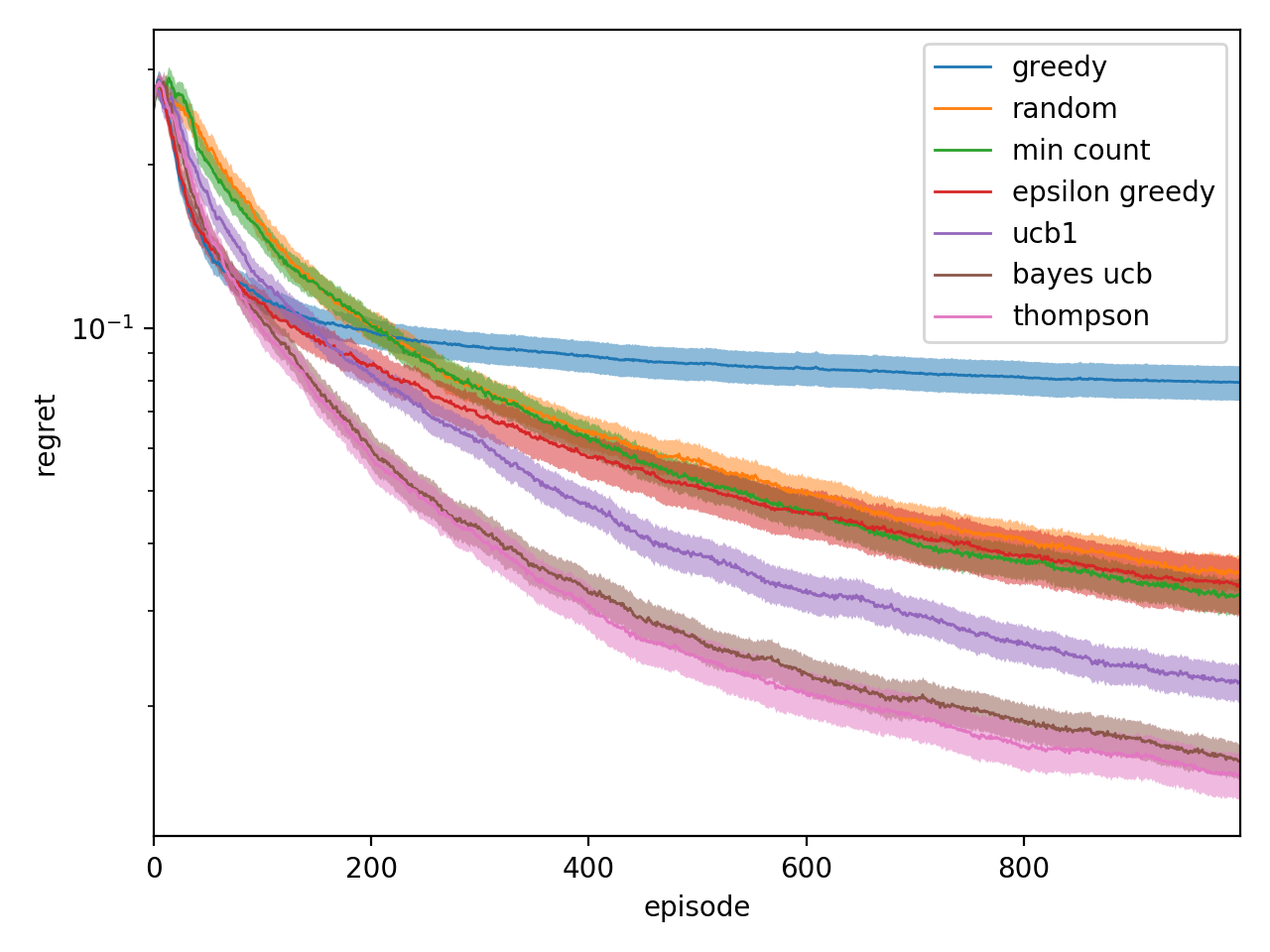}
\caption{Performance of each exploration method on a normal-form game with 10 actions for Player 1, 2 actions for Player 2, 1000 trials per method, and 100 belief samples.}
\label{fig:nfg1}
\end{figure}

In all experiments, the number of strategies sampled from the strategy spaces of each player that is used to estimate the UCBs and LCBs of the Bayes-UCB exploration strategy was 100.
Figure~\ref{fig:nfg1} shows the mean exploitability (with bands of 0.1 standard deviations) in a normal-form game that was randomly generated from the Jeffreys prior. The relative performance of the methods was the same across several generated games (not shown due to limited space).
\vspace{-.05in}

\begin{figure}[!ht]
\centering
\noindent \includegraphics[width=0.8\columnwidth]{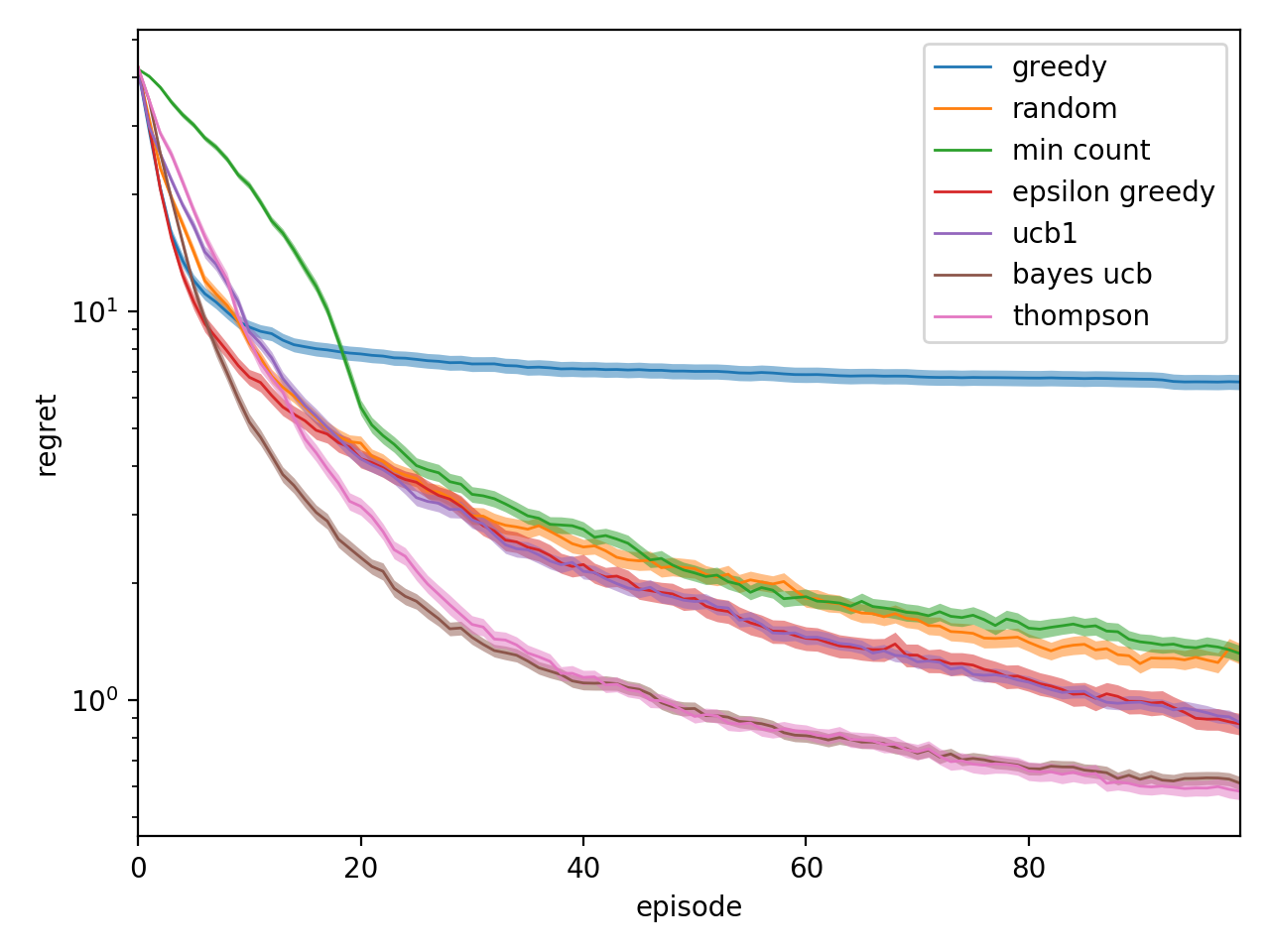}
\caption{Performance of each exploration method on a stochastic game with 10 states, 10 actions for Player 1, 2 actions for Player 2, a discount factor of 0.99, 100 trials per method, 100 belief samples, and 100 exploration steps per episode.}
\label{fig:sg1}
\end{figure}

Figure \ref{fig:sg1} shows performance on a stochastic game that was randomly generated from the Jeffreys prior. That is, the reward and transition probabilities for each action profile are sampled from the beta and Dirichlet distribution, respectively, with concentration parameters \(\frac{1}{2}\). The relative performance of the methods was the same across several generated games (not shown due to limited space).
\vspace{-.05in}

As expected, the greedy exploration method, while initially performing well, gets stuck at positive regret and is eventually outpaced by all the other methods. The random and mincount methods attain a similar level of performance. The epsilon-greedy method (in this case with \(\varepsilon = 0.1\)) attains better performance and, unlike the purely-greedy method, does not get stuck at a positive exploitability. UCB1 performed even better. The best-performing methods were Bayes-UCB and Thompson; this is not that surprising since the analogs of these strategies for the cumulative-regret version of the multi-armed bandit problem are known to perform well in that setting. It is also known that regret bounds for UCB can be converted into Bayesian regret bounds for Thompson sampling~\cite{Russo2014}. The latter has an advantage in terms of computational efficiency because it only needs to sample and solve one belief rather than many.

\section{Conclusions and future research}

We investigated the increasingly important and common game-solving setting where we do not have an explicit description of the game but only oracle access to it through game play. During a limited-duration learning phase, the algorithm can control the actions of both players in order to try to learn the game and how to play it well. After that, the algorithm has to produce a strategy that has low exploitability. 

We generalized the typical exploration strategies proposed for single-agent settings to normal-form and stochastic games. 
We proposed using the distribution of state-action value functions induced by a belief distribution over possible environments. 
The exploration strategies we evaluated were a strategy that chooses random action profiles, a strategy that chooses action profiles which have been explored the least, a greedy strategy that samples action profiles induced by the recommended strategies of each player, a Thompson-sampling-like strategy that samples an environment from its belief distribution and then samples an action profile from the resulting Nash equilibrium, and two UCB-like strategies that sample action profiles induced by optimistic strategies for each player. Bayes-UCB and Thompson sampling performed clearly the best.

In future work, we would like to see this research extended to nonzero-sum games and to more than two players. We would also like to see it extended to environments with large and/or continuous state spaces, where function approximation is required. Our exploration strategies suggest the following  approach for such settings. Maintain an ensemble of randomly-initialized neural networks that together represent the agent's belief distribution over environments.
Each network takes as input a state and action profile, and outputs the expected reward and next-state distribution, thus capturing the transition and reward structure of a hallucinated environment. Each network is repeatedly trained on the tuples of states, action profiles, rewards, and next states the agent has observed over the course of exploration.
In parallel, one trains separate Q-value networks on each of these environment networks using, for example, the value iteration algorithm that repeatedly applies Equation~\ref{eq:valiter} to randomly-sampled observation tuples that are queried from the hallucinated environment. This yields an ensemble of Q functions, one for each possible environment in the agent's belief distribution, which is what our exploration strategies use.

\section*{Acknowledgements}

This material is based on work supported by the National Science Foundation under grants IIS-1718457, IIS-1617590, IIS-1901403, and CCF-1733556, and the ARO under award W911NF-17-1-0082. The views expressed here are those of the authors and might not reflect those of their institutions or funding agencies.

\bibliographystyle{plain}
\bibliography{references}

\begin{thebibliography}{10}

\bibitem{Agrawal2017}
Shipra Agrawal and Randy Jia.
\newblock Optimistic posterior sampling for reinforcement learning: worst-case
  regret bounds.
\newblock In {\em Advances in Neural Information Processing Systems 30}, 2017.

\bibitem{Auer2002}
Peter Auer, Nicolò Cesa-Bianchi, and Paul Fischer.
\newblock Finite-time analysis of the multiarmed bandit problem.
\newblock {\em Machine Learning}, 2002.

\bibitem{Bellemare2017}
Marc~G. Bellemare, Will Dabney, and Rémi Munos.
\newblock A distributional perspective on reinforcement learning.
\newblock In {\em Proceedings of the 34th International Conference on Machine
  Learning}, 2017.

\bibitem{Berner2019}
Christopher Berner, Greg Brockman, Brooke Chan, Vicki Cheung, Przemysław
  Dębiak, Christy Dennison, David Farhi, Quirin Fischer, Shariq Hashme, Chris
  Hesse, Rafal Józefowicz, Scott Gray, Catherine Olsson, Jakub Pachocki,
  Michael Petrov, Henrique~Pondé de~Oliveira~Pinto, Jonathan Raiman, Tim
  Salimans, Jeremy Schlatter, Jonas Schneider, Szymon Sidor, Ilya Sutskever,
  Jie Tang, Filip Wolski, and Susan Zhang.
\newblock Dota 2 with large scale deep reinforcement learning, 2019.

\bibitem{Bubeck2009}
Sébastien Bubeck, Rémi Munos, and Gilles Stoltz.
\newblock Pure exploration in multi-armed bandits problems.
\newblock In {\em Algorithmic Learning Theory}, 2009.

\bibitem{Casgrain2019}
Philippe Casgrain, Brian Ning, and Sebastian Jaimungal.
\newblock Deep q-learning for nash equilibria: Nash-dqn, 2019.

\bibitem{Chen2017}
Richard~Y. Chen, Szymon Sidor, Pieter Abbeel, and John Schulman.
\newblock Ucb exploration via q-ensembles, 2017.

\bibitem{Claus1998}
Caroline Claus and Craig Boutilier.
\newblock The dynamics of reinforcement learning in cooperative multiagent
  systems.
\newblock In {\em Proceedings of the 15th National/10th Conference on
  Artificial Intelligence/Innovative Applications of Artificial Intelligence},
  1998.

\bibitem{Dearden1998}
Richard Dearden, Nir Friedman, and Stuart Russell.
\newblock Bayesian q-learning.
\newblock In {\em Proceedings of the 15th National/10th Conference on
  Artificial Intelligence/Innovative Applications of Artificial Intelligence},
  1998.

\bibitem{Ganzfried2009}
Sam Ganzfried and Tuomas~W. Sandholm.
\newblock Computing equilibria in multiplayer stochastic games of imperfect
  information.
\newblock In {\em Proceedings of the 21st International Joint Conference on
  Artificial Intelligence}, 2009.

\bibitem{Garivier2016}
Aurélien Garivier, Emilie Kaufmann, and Wouter~M. Koolen.
\newblock Maximin action identification: a new bandit framework for games.
\newblock In {\em 29th Annual Conference on Learning Theory}, 2016.

\bibitem{Hoffman1966}
Alan~J. Hoffman and Richard~M. Karp.
\newblock On nonterminating stochastic games.
\newblock {\em Management Science}, 1966.

\bibitem{Hu2003}
Junling Hu and Michael~P. Wellman.
\newblock Nash q-learning for general-sum stochastic games.
\newblock {\em Journal of Machine Learning Research}, 2003.

\bibitem{Jeffreys1946}
Harold Jeffreys.
\newblock An invariant form for the prior probability in estimation problems.
\newblock {\em Proceedings of the Royal Society of London. Series A,
  Mathematical and Physical Sciences}, 1946.

\bibitem{Kaufmann2012}
Emilie Kaufmann, Olivier Cappe, and Aurelien Garivier.
\newblock On bayesian upper confidence bounds for bandit problems.
\newblock In {\em Proceedings of the 15th International Conference on
  Artificial Intelligence and Statistics}, 2012.

\bibitem{Littman1994}
Michael~L. Littman.
\newblock Markov games as a framework for multi-agent reinforcement learning.
\newblock In {\em 11th International Conference on Machine Learning}, 1994.

\bibitem{Marchesi2019}
Alberto Marchesi, Francesco Trovò, and Nicola Gatti.
\newblock Learning probably approximately correct maximin strategies in
  simulation-based games with infinite strategy spaces, 2019.

\bibitem{Mavrin2019}
Borislav Mavrin, Hengshuai Yao, Linglong Kong, Kaiwen Wu, and Yaoliang Yu.
\newblock Distributional reinforcement learning for efficient exploration.
\newblock In {\em Proceedings of the 36th International Conference on Machine
  Learning}, 2019.

\bibitem{Mnih2015}
Volodymyr Mnih, Koray Kavukcuoglu, David Silver, Andrei~A. Rusu, Joel Veness,
  Marc~G. Bellemare, Alex Graves, Martin Riedmiller, Andreas~K. Fidjeland,
  Georg Ostrovski, Stig Petersen, Charles Beattie, Amir Sadik, Ioannis
  Antonoglou, Helen King, Dharshan Kumaran, Daan Wierstra, Shane Legg, and
  Demis Hassabis.
\newblock Human-level control through deep reinforcement learning.
\newblock {\em Nature}, 2015.

\bibitem{Donoghue2018}
Brendan O'Donoghue, Ian Osband, Rémi Munos, and Volodymyr Mnih.
\newblock The uncertainty bellman equation and exploration.
\newblock In {\em Proceedings of the 35th International Conference on Machine
  Learning}, 2018.

\bibitem{Ortega2010}
Pedro~A. Ortega and Daniel~A. Braun.
\newblock A minimum relative entropy principle for learning and acting.
\newblock {\em Journal of Artificial Intelligence Research}, 2010.

\bibitem{Osband2016}
Ian Osband, Charles Blundell, Alexander Pritzel, and Benjamin~Van Roy.
\newblock Deep exploration via bootstrapped dqn.
\newblock In {\em Advances in Neural Information Processing Systems 29}, 2016.

\bibitem{Osband2017}
Ian Osband and Benjamin~Van Roy.
\newblock Why is posterior sampling better than optimism for reinforcement
  learning?
\newblock In {\em Proceedings of the 34th International Conference on Machine
  Learning}, 2017.

\bibitem{Osband2019}
Ian Osband, Benjamin~Van Roy, Daniel~J. Russo, and Zheng Wen.
\newblock Deep exploration via randomized value functions.
\newblock {\em Journal of Machine Learning Research}, 2019.

\bibitem{Russo2014}
Daniel Russo and Benjamin~Van Roy.
\newblock Learning to optimize via posterior sampling.
\newblock {\em Mathematics of Operations Research}, 2014.

\bibitem{Russo2018}
Daniel~J. Russo, Benjamin~Van Roy, Abbas Kazerouni, Ian Osband, and Zheng Wen.
\newblock A tutorial on thompson sampling.
\newblock {\em Foundations and Trends in Machine Learning}, 2018.

\bibitem{Sandholm1996}
Tuomas~W. Sandholm and Robert~H. Crites.
\newblock Multiagent reinforcement learning in the iterated prisoner's dilemma.
\newblock {\em Biosystems}, 1996.

\bibitem{Shapley1953}
Lloyd~S. Shapley.
\newblock Stochastic games.
\newblock {\em Proceedings of the National Academy of Sciences}, 1953.

\bibitem{Sokota2019}
Samuel Sokota, Caleb Ho, and Bryce Wiedenbeck.
\newblock Learning deviation payoffs in simulation-based games.
\newblock In {\em Proceedings of the AAAI Conference on Artificial
  Intelligence}, 2019.

\bibitem{Thompson1933}
William~R. Thompson.
\newblock On the likelihood that one unknown probability exceeds another in
  view of the evidence of two samples.
\newblock {\em Biometrika}, 1933.

\bibitem{Vinyals2019}
Oriol Vinyals, Igor Babuschkin, Wojciech~M. Czarnecki, Michaël Mathieu, Andrew
  Dudzik, Junyoung Chung, David~H. Choi, Richard Powell, Timo Ewalds, Petko
  Georgiev, Junhyuk Oh, Dan Horgan, Manuel Kroiss, Ivo Danihelka, Aja Huang,
  Laurent Sifre, Trevor Cai, John~P. Agapiou, Max Jaderberg, Alexander~S.
  Vezhnevets, R{\'{e}}mi Leblond, Tobias Pohlen, Valentin Dalibard, David
  Budden, Yury Sulsky, James Molloy, Tom~L. Paine, Caglar Gulcehre, Ziyu Wang,
  Tobias Pfaff, Yuhuai Wu, Roman Ring, Dani Yogatama, Dario W\"{u}nsch, Katrina
  McKinney, Oliver Smith, Tom Schaul, Timothy Lillicrap, Koray Kavukcuoglu,
  Demis Hassabis, Chris Apps, and David Silver.
\newblock Grandmaster level in starcraft ii using multi-agent reinforcement
  learning.
\newblock {\em Nature}, 2019.

\bibitem{Neumann1928}
John von Neumann.
\newblock Zur theorie der gesellschaftsspiele.
\newblock {\em Mathematische Annalen}, 1928.

\bibitem{Vorobeychik2009}
Yevgeniy Vorobeychik and Michael~P. Wellman.
\newblock Strategic analysis with simulation-based games.
\newblock In {\em Proceedings of the 2009 Winter Simulation Conference}, 2009.

\bibitem{Wang2002}
Xiaofeng Wang and Tuomas~W. Sandholm.
\newblock Reinforcement learning to play an optimal nash equilibrium in team
  markov games.
\newblock In {\em Proceedings of the 15th International Conference on Neural
  Information Processing Systems}, 2002.

\bibitem{Watkins1992}
Christopher J. C.~H. Watkins and Peter Dayan.
\newblock Q-learning.
\newblock {\em Machine Learning}, 1992.

\end{thebibliography}

\end{document}